\pdfoutput=1
\documentclass[11pt]{article}

\usepackage[]{naacl2021}

\usepackage{times}
\usepackage{latexsym}
\usepackage{graphicx}
\usepackage{amsmath}
\usepackage{amssymb}
\usepackage{subfig}

\usepackage[T1]{fontenc}
\usepackage[utf8]{inputenc}

\usepackage{microtype}

\title{OodGAN: Generative Adversarial Network for Out-of-Domain Data Generation}

\author{Petr Marek\thanks{\hspace{0.2cm}Research conducted during an internship at Amazon Alexa AI} \\ Czech Technical University in Prague \\ Prague, Czech Republic \\ \texttt{marekp17@fel.cvut.cz}
  \And
  Vishal Ishwar Naik \\ Amazon Alexa AI \\ Sunnyvale, California \\ \texttt{naikvish@amazon.com} 
  \AND
  Vincent Auvray \\ Amazon Alexa AI \\ Sunnyvale, California \\ \texttt{vauvray@amazon.de}
  \And
  Anuj Goyal \\ Amazon Alexa AI \\ Sunnyvale, California \\ \texttt{anujgoya@amazon.com}}

\begin{document}
\maketitle
\begin{abstract}
Detecting an Out-of-Domain (OOD) utterance is crucial for a robust dialog system. Most dialog systems are trained on a pool of annotated OOD data to achieve this goal. However, collecting the annotated OOD data for a given domain is an expensive process. To mitigate this issue, previous works have proposed generative adversarial networks (GAN) based models to generate OOD data for a given domain automatically. However, these proposed models do not work directly with the text. They work with the text's latent space instead, enforcing these models to include components responsible for encoding text into latent space and decoding it back, such as auto-encoder. These components increase the model complexity, making it difficult to train.

We propose OodGAN, a sequential generative adversarial network (SeqGAN) based model for OOD data generation. Our proposed model works directly on the text and hence eliminates the need to include an auto-encoder. OOD data generated using OodGAN model outperforms state-of-the-art in OOD detection metrics for ROSTD (67\% relative improvement in FPR 0.95) and OSQ datasets (28\% relative improvement in FPR~0.95) \citep{zheng2020out}. 
%OOD data generated using the OodGAN model for ROSTD dataset \citep{gangal2019likelihood} outperforms the state of art OOD detection metrics and works better than \citet{zheng2020out} for OSQ dataset.
\end{abstract}

\section{Introduction}

OOD detection is an essential task in AI voice assistants like Alexa, Siri, or Google Assistant. The task is to recognize whether a given user utterance belongs to the in-domain (IND) distribution or not. Users usually do not know the limitations of a voice application and assign requests which the system can not act upon. These requests are referred to as OOD since these do not belong to the application's domain. Voice assistants should be able to handle OOD utterances robustly by not taking unintended action or giving wrong or nonsensical responses leading to a poor user experience.

Intent classification (IC) is one of the main tasks in a conversational system that selects the best intent given a user input. IC can be extended to support OOD detection in two different ways. The first one is to add OOD as another intent to the IC model, but this requires annotated OOD data for training. The second method is to use a threshold on the classifier's output probability distribution during the runtime. This method does not require OOD data for training necessarily. Nevertheless, it proves difficult to select the threshold in practice without it.

The state-of-the-art IC algorithms are trained using neural networks to produce probability distribution over output classes and use cross-entropy loss. However, \citet{lakshminarayanan2017simple}, and \citet{guo2017calibration} pointed out that the neural network classifier tends to be overconfident in its classification. This means that the classifier tends to assign a high probability for one class, even when the example was not seen in the training phase. Thus, such a classifier cannot correctly recognize if an example belongs to an IND or OOD distribution during runtime with any reasonable threshold value. In this paper, we focus on improving the performance of the threshold-based OOD detection method with the help of generated OOD data.

\citet{zheng2020out} proposed to use negative entropy as an additional loss for the classification task in a neural network. The negative entropy loss trains the network to flatten the produced probability distribution as opposed to cross-entropy, which teaches the network to maximize the correct class probability. Thus, the idea is to apply cross-entropy loss on IND data and negative entropy loss on OOD data. The result is that IND data receives a high probability for the correct class, and OOD data receives low probabilities for all classes. Thanks to this fact, we can select a reasonable threshold on the output probability that will classify both IND and OOD data correctly. We need OOD data to train models in this way. However, the collection of OOD data is a manual and expensive process.

The IND data forms a small distribution cluster in the space of vector text representation. In principle, the rest of that space is covered by OOD data. Also, in real-world scenarios, most OOD data share patterns with IND data. Nevertheless, \citet{zheng2020out} demonstrated that training IC model with OOD data that are just outside IND distribution should be sufficient to handle most of the OOD requests during runtime. 

In this paper, we propose a novel OOD data generation model OodGAN, which is an extension of SeqGAN \cite{yu2017seqgan}. We use GAN to generate OOD data that share the same patterns as IND and are very close to IND distribution. 
%We have tested our proposed model on two datasets, ROSTD and OSQ. The proposed model outperforms the-state-of-the-art OOD metrics on the ROSTD dataset, while the OSQ dataset results are promising.

Our proposed model aims to be deployed to Natural Language Understanding (NLU) frameworks offered by popular voice assistants like Amazon Alexa and Google Assistant. These NLU frameworks are offered to third-party developers to create voice applications. 
%(called ``skills'' in Alexa and ``actions'' in Google Home). 
Third-party developers can define any number of IND intents and provide sample utterances for each to build voice applications. These voice applications should recognize OOD requests during run time without additional developer effort to provide OOD training data. The proposed model can be deployed in a NLU framework to generate application-specific OOD data that the IC model can use during training to recognize OOD requests robustly and improve the end-user experience.

%Our main contributions are:
%\begin{itemize}
%    \item We show shortcomings of the OOD data generation model proposed by \citet{zheng2020out}.
%    \item We propose a novel and simpler OOD data generation model OodGAN. It works with a sequence of words directly unlike the previously proposed models, which work on latent space represented by auto-encoder. Our model eliminates the need for the auto-encoder, which reduces the overall size of the model.
%    \item We evaluate our model on the ROSTD and OSQ datasets, and we show that OOD examples generated by OodGAN achieved state-of-the-art results.
%\end{itemize}

Our main contributions are: 

\textbf{(1)} We propose a novel and simple OOD data generation model OodGAN that improves on the model proposed by \citet{zheng2020out}. It works with a sequence of words directly unlike the previously proposed models, which work on latent space represented by auto-encoder. Our model eliminates the need for the auto-encoder, which reduces the overall size of the model.

\textbf{(2)} We evaluate our model on the ROSTD and OSQ datasets, and we show that OOD examples generated by OodGAN achieved state-of-the-art results.

\section{Related Work}

There are three research areas relevant to our work: OOD detection, text generation and OOD generation. 

\subsubsection*{Out-of-Domain Detection}

%\citet{larson2019evaluation} introduced a dataset for intent classification that includes queries that are IND as well as OOD. The OOD queries share similar patterns with IND queries but are not supported in the domain. They propose three baseline approaches for the task of predicting whether an example is OOD or not. The first method considers OOD examples as an additional intent class. The second method uses a threshold on the classifier's probability estimates. The third uses a two-stage process where the classifier first classifies a query as IND or OOD and then classifies it into one of the 150 intents if classified as IND. The disadvantage of all these baseline methods is their reliance on OOD training data, which is not scalable to collect.

\citet{larson2019evaluation} introduced a dataset for intent classification that includes OOD queries. They propose three baseline approaches for OOD detection that rely on OOD training data. 
\citet{gangal2019likelihood} created a ROSTD dataset and explored likelihood ratio based approaches. %Specifically, they reformulated and applied these approaches to natural language inputs. 
%They also proposed learning a generative classifier and computing a marginal likelihood (ratio) for OOD detection. This approach allows using a principled likelihood while at the same time exploiting training-time labels.
\citet{lee2019contextual} proposed an OOD detection method that does not require OOD data by utilizing counterfeit OOD turns in the context of a dialog. 
%They also show that independent decisions without consideration of dialogue context lead to suboptimal performance.
%\citet{lee2017training} developed a novel training method for classifiers to discriminate IND and OOD examples better. They suggest two additional terms added to the original loss (cross-entropy). The first one forces examples from out-of-distribution less confident by the classifier. The second one is for (implicitly) generating the most effective training examples for the first one. In essence, their method jointly trains both classification and generative neural networks for OOD distribution.
%\citet{lee2017training} developed a novel training method for OOD detection. They suggest two additional terms added to the original cross-entropy loss.
%\citet{ryu2018out} proposed an OOD detection system that uses only IND sentences to build a generative adversarial network in which the discriminator generates low scores for OOD sentences. To improve basic GANs, they apply feature matching loss in the discriminator, use domain-category analysis as an additional task in the discriminator, and remove the generator's biases.
\citet{ryu2018out} proposed an OOD detection system that uses only IND sentences to build a generative adversarial network in which the discriminator generates low scores for OOD sentences.

\subsubsection*{Text Generation}
%\textcolor{red}{\citet{donahue2018adversarial} proposed a two-step solution to text generation. The first step is to utilize an auto-encoder to learn a low-dimensional representation of sentences. A GAN is then trained to generate vectors in this space, decoded to realistic utterances in the second step. Nevertheless, this approach needs a lot of training data for the auto-encoder to learn how to reconstruct utterances.}
\citet{donahue2018adversarial} proposed a two-step solution to text generation using auto-encoder and GAN that works with a low-dimensional representation of sentences.
%\citet{yu2017seqgan} proposed a sequence generation framework called SeqGAN. Modeling the data generator as a stochastic policy in reinforcement learning, SeqGAN bypasses the generator differentiation problem by directly performing gradient policy update. The reinforcement learning reward signal comes from the GAN discriminator judged on a complete sequence and is passed back to the intermediate state-action steps using a Monte Carlo search.
\citet{yu2017seqgan} proposed a sequence generation framework SeqGAN that works directly on the text and hence eliminates the need for an auto-encoder.

\subsubsection*{Out-of-Domain Data Generation}

\citet{zheng2020out} proposed a GAN based model to generate pseudo-OOD examples that are akin to IND input utterances. The model uses a denoising auto-encoder that is trained to map an input example into a latent code. The functions of the auto-encoder's parts are the following. The encoder learns to create a latent representation of the examples. The decoder learns to convert the vector of the latent representation into text. The model's generator produces vectors in the latent space. The discriminator evaluates the closeness of latent space vectors generated by the generator to real latent space vectors created by the encoder. Discriminator sends a training signal to the generator to force it to generate indistinguishable vectors from vectors encoded by the encoder. An auxiliary classifier trained on IND examples is introduced to force the generator to generate latent code belonging to OOD. 
%The decoder decodes the resulting generated latent code vectors to text form. 
The resulting utterances share patterns with IND examples but belong to OOD.

\section{Generative Adversarial Networks for Out-of-Domain Data Generation}

\begin{figure}[!t]
\centering
\includegraphics[width=2.5in]{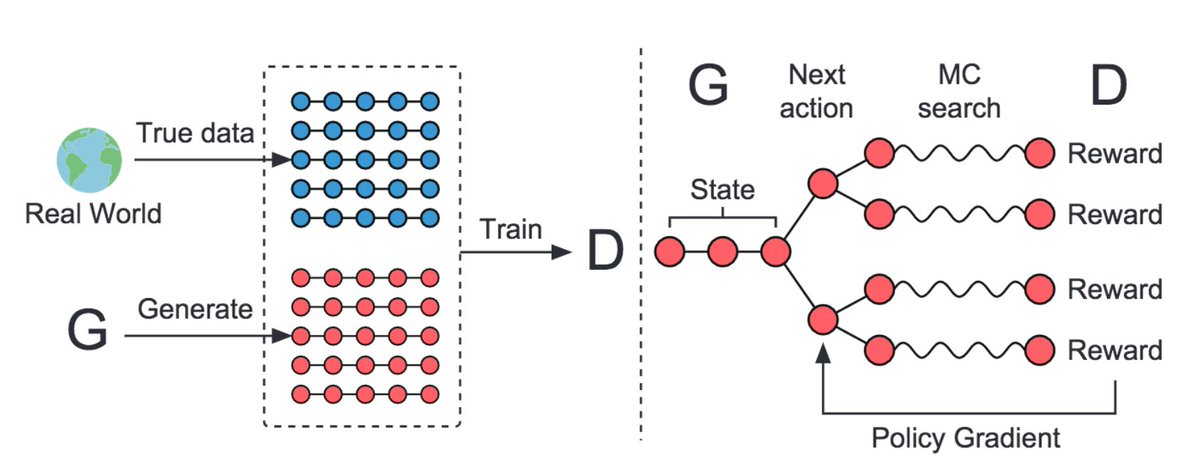}
\caption{The illustration of SeqGAN \citep{yu2017seqgan}. Left: Discriminator~\(D\) is trained over
the real data and the data generated by generator~\(G\). Right: Generator is trained by policy gradient where the final reward signal is provided by  the discriminator and is passed back to the intermediate action value via Monte Carlo search.}
\label{SegGAN}
\end{figure}

\begin{figure}[!t]
\centering
\includegraphics[width=2.5in]{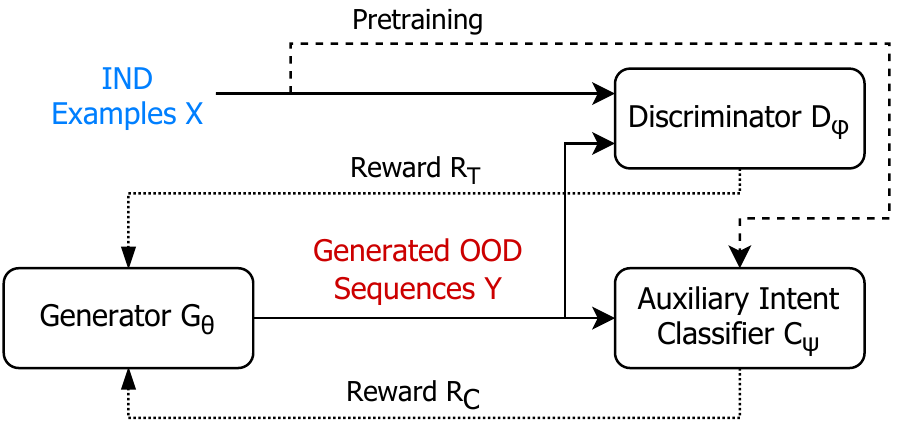}
\caption{
%Overall architecture of the OodGAN. The model consists of Generator \(G_\theta\), Discriminator \(D_\phi\) and Auxiliary Intent Classifier \(C_\psi\). \(C_\psi\) is pretrained to recognize intent classes $z_{1...m}$ for IND examples $X$. \(D_\phi\) is trained to distinguish between $X$ and generated OOD examples $Y$ during adversarial training. %\(G_\theta\) is trained by REINFORCE algorithm \citep{williams1992simple} during adversarial training to generate OOD sequences $Y$. Generated $Y$ is evaluated by \(D_\phi\) and \(C_\psi\), and corresponding rewards \(R_T\) and \(R_C\) are given to \(G_\theta\). \(R_T\) rewards \(G_\theta\) for generating sequences similar to IND examples $X$. \(R_C\) rewards \(G_\theta\) for generating sequences not belonging into any IND intent class $z_{1...m}$.}
The overall architecture of the OodGAN. \(C_\psi\) is pretrained to recognize intent classes for IND examples. \(D_\phi\) is trained to distinguish between IND and generated OOD examples during adversarial training.
\(G_\theta\) is trained by the REINFORCE algorithm during adversarial training to generate OOD sequences. The training is guided by rewards originating in \(C_\psi\) and \(D_\phi\).}
\label{OodGAN}
\end{figure}

%The problem of OOD data generation is denoted as follows. Given the dataset of IND examples $X_{1...n}$ belonging to classes $z_{1...m}$ where $n>m$, we want to generate examples $Y_{1...p}$ sharing patterns with the IND examples $X_{1...n}$ but at the same time not belonging to any class $z_{1...m}$. We use the SeqGAN model proposed by \citet{yu2017seqgan} as a starting point.

\subsection{SeqGAN}
The SeqGAN model proposed by \citet{yu2017seqgan} is a starting point for the proposed OodGAN.
SeqGAN is a sequence generation framework illustrated in \autoref{SegGAN}. \citet{yu2017seqgan} denote the problem of sequence generation as follows. Given a dataset of real-world structured sequences, train a \(\theta\)-parameterized generative model \(G_\theta\) to produce a sequence \(Y_{1:T} = (y_1,...,y_t,...y_T), y_t \in Y\) where \(Y\) is the vocabulary of candidate tokens. They apply reinforcement learning to this problem. In timestep \(t\), the state \(s\) is the current produced tokens \((y_1, ... , y_{t-1})\) and the action \(a\) is the next token \(y_t\) to select.

They propose to additionally train a \(\phi\)-parameterized discriminative model \(D_\phi\) that provides guidance for improving generator \(G_\theta\). \(D_\phi\) produces a probability \(D_\phi(Y_{1:T})\) representing the probability of \(Y_{1:T}\) being a real sequence vs. a generated one. The discriminative model \(D_\phi\) is trained with real sequence data, labeled as positive examples, and synthetic sequences from the generative model \(G_\theta\), labeled as negative examples.

SeqGAN uses the REINFORCE algorithm \citep{williams1992simple} to train generative model \(G_\theta\).
Parameters of generative model \(G_\theta\) are updated at the same time by a policy gradient and Monte Carlo search based on the expected end reward received from the discriminative model \(D_\phi\) for the generated sequence. The reward is represented by a likelihood that the generated sequence will fool the discriminative model \(D_\phi\).
%The objective of the generator is to generate a sequence from the start state \(s_0\) to maximize its expected reward:
%\begin{equation*}
%\resizebox{\hsize}{!}{$J(\theta) = \mathbb{E}[R_T|s_0,\theta] = \sum_{y_1 \in Y}G_\theta(y_1|s_0) \cdot Q_{D_\phi}^{G_\theta}(s_0, y_1),$}
%\end{equation*}
%where \(R_T\) is a reward for a whole sequence from the discriminator \(D_\phi\), \(G_\theta(y_t|Y_{1:t-1})\) is generator model policy, and \(Q_{D_\phi}^{G_\theta}(s, a)\) is the action-value function of a sequence. The action-value function returns expected accumulative reward starting from state \(s\), taking action \(a\), and then following policy \(G_\theta\). This equation's meaning is that the generator's goal is to generate a sequence from an initial state that would fool the discriminator into considering it as real.
Thus the generator's goal is to generate a sequence that would fool the discriminator into considering it as real.

\subsection{OodGAN}

We propose OodGAN based on SeqGAN. There are two benefits of SeqGAN for our task of OOD data generation. SeqGAN produces sequences similar to the training data, and it works directly on input sequence unlike earlier model \citep{zheng2020out}, which works on latent space. Eliminating the auto-encoder responsible for converting a sequence of words into latent space reduces the overall model size. Also, our experiments with \citet{zheng2020out} show a degradation in the overall performance due to the auto-encoder component (see the Results section for details).

Since our task is to generate OOD data, we have the additional criterion that generated sequences should be close to the training IND sequences. However, we also want them not to belong to any IND intent class. We propose the OodGAN to achieve the two criteria.

The main difference between SeqGAN and OodGAN is the introduction of an auxiliary intent classifier. The auxiliary intent classifier \(C_\psi\) estimates the probability \(C_\psi(z_i|Y)\) of example \(Y\) belonging into intent class \(z_i\).
%\begin{equation*}
%C_\psi(z_i|Y)
%\end{equation*}
The task of the auxiliary intent classifier is to produce an additional reward signal. The reward signal guides the generator to produce a sequence not belonging to any IND intent class. The reward \(R_{C_\psi}\) coming from the auxiliary intent classifier for each generated example is defined as Shannon's Entropy \(R_{C_\psi}= -\sum_{i=1}^{m} C_\psi(z_i|Y)\cdot log(C_\psi(z_i|Y))\),
%\begin{equation*}
%R_C= -\sum_{i=1}^{m} C_\psi(z_i|Y)\cdot log(C_\psi(z_i|Y)),
%\end{equation*}
where \(m\) is the number of IND intent classes. 
The intuition for using Shannon's Entropy is that we want to reward a generator for producing examples for which the auxiliary intent classifier cannot clearly assign one of IND classes. In other words, the auxiliary classifier should assign a nearly uniform distribution across all intent classes for a good generated example. The generator obtains a high reward for such examples because the uniform distribution has the highest Shannon's Entropy.

We train the auxiliary intent classifier to predict one of the classes $z_{1...m}$ for each training IND example $X_{1...n}$ during the pre-training step. We do not have to retrain it during adversarial training because IND intent classes' distribution does not change.

%The goal of the generator is to generate a sequence from the start state \(s_0\) that maximizes the expected sum of rewards from discriminator \(D_\phi\) and auxiliary intent classifier \(C_\psi\):
%\begin{equation*}
%\resizebox{\hsize}{!}{$\begin{aligned}
%& J(\theta) = \mathbb{E}[R_T+R_C|s_0,\theta] = \sum_{y_1 \in Y}G_\theta(y_1|s_0) \cdot \\
%&\cdot Q_{D_\phi+C_\psi}^{G_\theta}(s_0, y_1).\end{aligned}$}
%\end{equation*}

The goal of the generator is to generate a sequence that maximizes the expected sum of rewards from discriminator \(D_\phi\) (the estimated probability of the sequence being real), and auxiliary intent classifier \(C_\psi\) (Shannon's Entropy calculated using estimated probabilities of sequence belonging to IND intent classes by auxiliary intent classifier).

Empirically, we evaluated different training strategies. We found that optimizing generator \(G\) using only the discriminator's reward first, followed by using only the auxiliary intent classifier reward, and then repeating the process for each training batch produced the most stable results. This worked better than summing up the rewards from the discriminator and auxiliary intent classifier. When we tried summing up the two rewards, we noticed that the generator tended to collapse into a state in which it generated a single sequence highly rewarded by the auxiliary intent classifier, even though this did not happen for all training runs. We observed this situation even when we normalized rewards to a value between 0 and 1. 
%The normalization of Shanon's entropy is possible. We can calculate its maximal value since we know the number of IND intent classes \(m\).

We also observed that part of the examples generated by OodGAN is semantically similar to some IND training example or is generated multiple times. Examples that are identical or too close to IND examples are problematic and confuse the OOD classifier. Duplicated examples do not represent the OOD distribution effectively. For those reasons, we removed with an automatic filter the generated OOD examples that are identical or similar to IND examples or that are generated repeatedly.

To summarize, OodGAN's training procedure has the following steps.

\textbf{(1)~Train Auxiliary classifier:} First train auxiliary classifier to predict the classes $z_{1...m}$ for IND data $X_{1...n}$ until convergence. 

\textbf{(2)~Train Generator as Language Model:} Next, train the generator on the IND data $X_{1...n}$  as a language model until it converges. Thanks to this step, it is easier for the generator to fool the discriminator from the start of the adversarial training.

\textbf{(3)~Train Discriminator:} Generate adversarial examples from the generator. This training step helps the discriminator to provide a useful reward signal from the start of adversarial training.

\textbf{(4)~Adversarial Training:} Perform adversarial training of generator and discriminator. There are three optimization steps for each training batch. First, optimize the generator using reward from discriminator as proposed by \citet{yu2017seqgan}. Next, optimize the generator using a reward from the auxiliary classifier. Lastly, optimize the discriminator.

\section{Experiments}

\subsection{Datasets}

We conducted experiments on ROSTD \citep{gangal2019likelihood} and OSQ \citep{larson2019evaluation} datasets. 
%OOD examples in the OSQ are more similar to IND than in ROSTD dataset.
%OOD examples in the ROSTD dataset are far from IND, and in the OSQ dataset, OOD examples are very close to IND. 
%This difference is caused by the way how these two datasets were created.
%ROSTD contains 12 intents, and its OOD examples were selected from different domains not to share patterns with any IND examples. 
%OSQ contains 150 intents. OOD examples in the OSQ data share similar patterns with IND examples and are right outside the IND-OOD boundary. 

\begin{itemize}
    \item \textbf{ROSTD} contains three categories (alarm, reminder, and weather), each consisting of four intents. The dataset consists of 30,000 training, 4,000 validation and 8,000 testing IND examples. OOD examples were selected in a way that they do not belong to any category and do not share patterns with any IND examples. There are also no OOD examples in the training set of the dataset. The testing set contains 4,500 OOD examples.
    IND and OOD examples from ROSTD are listed in \autoref{tab:ROSTD-examples}.
    \item \textbf{OSQ} consists of 150 intents. The datases consists of 15,000 training, 3,000 validation and 4,500 testing IND examples. The dataset was created using Mechanical Turk. The turkers were given the name of the intent, and they were supposed to write intent examples fitting into the intent. The dataset authors manually went through examples and moved examples not fitting into the given intent class to the OOD class. In this way, OOD examples share the same patterns as IND examples.
    The OSQ dataset contains 100 training OOD examples. However, we decided not to use them for training due to the nature of our experiments. There are also 100 validation and 1,000 testing OOD examples.
    %IND and OOD examples from the OSQ dataset are listed in \autoref{tab:OSQ-examples}.
\end{itemize}

%OOD examples during runtime do not have fixed distribution. Some OOD examples are close to IND, and some OOD examples are far. We experimented with two different datasets for the reason that each of them tests a different described situation.

\subsection{Evaluation Process}

%One way to evaluate the performance of OodGAN is to measure the quality of generated OOD data. We need a pool of annotated OOD data to compare with generated data to measure the quality. But, the distribution of OOD data is limitless, and hence it is unfair to compare generated OOD with any pool of annotated OOD data. However, we can use generated OOD data in the downstream task of OOD data detection and measure the change in OOD data detection metrics.

We evaluate the model on the downstream task of OOD data detection and measure the change in OOD data detection metrics.
We designed experiments in the following way. We train the OodGAN on IND training examples as a first step. Next, we generate the OOD examples using the trained model of OodGAN. We generate the same number of OOD examples as a number of IND examples in the training set. In a third step, we train the threshold-based OOD detection model using cross-entropy loss on training IND examples and negative entropy loss on generated OOD examples. In the last step, we evaluate both IND and OOD metrics.

\subsection{Metrics}

We evaluate the OodGAN by measuring metrics on the downstream task of OOD detection. We measure AUROC, AUPR, and FPR\emph{N} metrics \citep{Ren2019LikelihoodRF,Hendrycks2017ABF,Hendrycks2019DeepAD} to evaluate OodGAN's ability to generate OOD data that helps IC to distinguish IND and OOD input utterances. We treat OOD examples as the positive class.
%We measure AUROC, AUPR, and FPR\emph{N} metrics \citep{Ren2019LikelihoodRF,Hendrycks2017ABF,Hendrycks2019DeepAD} to evaluate OodGAN's ability to generate OOD data that helps IC to distinguish IND and OOD input utterances. We treat OOD examples as the positive class.
%We also measure IND accuracy that evaluates generated OOD data's influence on the IC's ability to recognize the intents of IND data correctly. 

\begin{itemize}
    \item \textbf{AUROC} The area under the receiver operating characteristic (ROC) curve. The score says the probability that a randomly selected OOD example will have a higher predicted probability of being an OOD than a randomly selected IND example. Higher AUROC score is better.
    \item \textbf{AUPR} The area under the precision-recall curve when OOD inputs are treated as positive samples. AUPR calculates the average precision score for all recall values. Intuitively, the higher the classification threshold we select, the more OOD will be classified as OOD. However, we risk that more IND will be classified as OOD. AUPR expresses this risk. Higher AUPR score is better.
   \item \textbf{FPR\emph{N}} The false-positive rate (FPR) when the true positive rate (TPR) is N\%. FPR\emph{N} metric is a practical value in real-world application since it evaluates an OOD detection performance at a particular threshold. Lower FPR\emph{N} means there is a smaller chance of IND examples triggering false alarm (IND getting classified as OOD) when the model's performance on OOD example is N\%. We report FPR when TPR is 0.95 and 0.90. Lower FPR\emph{N} score is better.
\end{itemize}
We consider FPR\emph{N} metric as the most practical value in real-world application since it evaluates an OOD detection performance at a particular threshold. Lower FPR\emph{N} means there is a smaller chance of IND examples triggering false alarm (IND getting classified as OOD) when the model correctly recognizes N\% of OOD examples.

We also measure IND accuracy that evaluates generated OOD data's influence on the IC's ability to recognize the intents of IND data correctly.
\begin{itemize}
    \item \textbf{IND accuracy} The percentage of IND data that have assigned correct intent label. We expect that generated OOD examples cannot improve the IC's ability to recognize intent labels for ID. However, generated OOD examples can degrade the IC's ability to recognize IND intents. Thus, we measure the IND accuracy to evaluate whether generated OOD negatively impacts the IC. Higher IND accuracy is better.
\end{itemize}

\subsection{Implementation}

We based our implementation on the Github repository\footnote{https://github.com/suragnair/seqGAN} of SeqGAN implemented in PyTorch. The generator is one layer GRU recurrent neural network trained using Adam optimizer with a learning rate set to 0.001. Input to the generator is embedded with fastText embeddings \citep{joulin2016bag} trained on Wikipedia. The generator uses negative log-likelihood loss during LM training and policy gradient loss during GAN training. The discriminator is a two-layer bidirectional GRU recurrent neural network with a tanh activation function. Adagrad optimization is used for training the discriminator with a learning rate set to 0.1 and binary cross-entropy loss is optimized. The auxiliary classifier uses the convolutional neural network proposed by \citet{kim2014convolutional}, which has filters of size 2, 3, 4, and 5, and for each size, there are 256 filters. We used the LeakyReLU activation function and 0.5 dropout in output dense layers. The auxiliary classifier is trained using the Adam optimizer with a learning rate set to 0.0001 and cross-entropy loss is optimized.

We show the comparison of number of parameters between OodGAN, SeqGAN, and \citet{zheng2020out} in \autoref{tab:numberOfParams}.
 
\begin{table}[h]
\centering
\begin{tabular}{l|l}
\hline
       & \# Parameters \\ \hline
\citet{zheng2020out}  & 7M            \\
SeqGAN \citep{yu2017seqgan} & 800k          \\
OodGAN & 2M            \\ \hline
\end{tabular}
\caption{Number of parameters}
\label{tab:numberOfParams}
\end{table}

%\subsection{Filtering Mechanism}

%During experiments, we observed that part of the examples generated by OodGAN is semantically similar to some IND training example or is generated multiple times. Examples that are equal or too close to IND examples are problematic and confuse the OOD classifier. Duplicated examples do not represent the OOD distribution effectively. For those reasons, we propose the Filtering Mechanism, which is responsible for removing generated OOD examples that are equal or similar to IND examples and are generated repeatedly. We designed two variants of Filtering Mechanism:
%\begin{itemize}
%    \item \textbf{Hard Filter} removes an example if it has been generated already or is equal to some training IND example.
%    \item \textbf{Soft Filter} removes an example if it has been generated already or if the auxiliary intent classifier predicts its class with a probability higher than the selected threshold. We experimented with different thresholds.  
%\end{itemize}

\section{Results}

%\subsection{Results on model proposed by \citet{zheng2020out}}

\subsection{Results on \citet{zheng2020out}}

We first conducted experiments to replicate results reported by \citet{zheng2020out} on the OSQ dataset. We created our implementation according to the paper's description because there is no publicly accessible implementation of their proposed model. We report results in \autoref{tab:OSQold-results}.

\begin{table}[]
\resizebox{\columnwidth}{!}{%
\begin{tabular}{@{}l|lllll@{}}
\hline
\multicolumn{1}{r|}{\begin{tabular}[c]{@{}r@{}}OSQ \\ \citep{larson2019evaluation} \end{tabular}} & AUROC $\uparrow$          & AUPR $\uparrow$           & \begin{tabular}[c]{@{}l@{}} FPR \\ 0.95 \end{tabular} $\downarrow$       & \begin{tabular}[c]{@{}l@{}} FPR \\ 0.90 \end{tabular} $\downarrow$        & \begin{tabular}[c]{@{}l@{}}IND \\ Acc.\end{tabular} $\uparrow$ \\ \hline
\begin{tabular}[c]{@{}l@{}}Results reported \\ by \citet{zheng2020out}\end{tabular}                        & 95.4          & 98.9          & 25.0           & 10.1           & 93.3                                               \\
\begin{tabular}[c]{@{}l@{}}Our implementation of \\ \citet{zheng2020out}\end{tabular}& 88.79          & 58.22 & 36.49          & 26.87                & 88.00  
\\ \hline
%Our Implementation of \citet{zheng2020out}              & 90.9          & 63.8 & 30.4          & 22.8                & 90.7                                                             \\ \hline
\end{tabular}
}
\caption{OOD detection performance on the OSQ dataset with model proposed by \citet{zheng2020out}}
\label{tab:OSQold-results}
\end{table}

%We could not reproduce the number reported by \citet{zheng2020out} even though we implemented the model as was described in the paper. However, we found a weak part of their architecture. Their GAN model used denoising auto-encoder to convert text data to latent space and back. It is an integral part of the model, enabling the generator to generate a fixed-sized vector representation of utterance in latent space. Decoder later converts this latent representation back into a text representation. Our measurements showed that the auto-encoder trained from scratch on the OSQ dataset performed poorly. The token accuracy of text reconstruction on the validation set was only 0.37\%.  The low performance of the auto-encoder is the reason why the generator generates meaningless sentences. Hence the performance of OOD detection trained on meaningless sentences suffers. We hypothesize that using transformer \citep{Lewis2020BARTDS} based auto-encoder would help solve this problem. We leave this for our future work. 

We could not reproduce the number reported by \citet{zheng2020out} even though we implemented the model as was described in the paper. The experiments showed that the denoising auto-encoder is a weak part of the architecture. Its token accuracy of text reconstruction on the validation set was only 0.37\%. Thus, the low performance of the auto-encoder is the reason why the generator generates poor quality examples.

\subsection{Results on proposed model OodGAN}

First, we want to compare OodGAN with baselines. We selected two baselines to evaluate improvements of our proposed OodGAN. Our baselines for the ROSTD dataset is our implementation of \citet{zheng2020out} and the work of \citet{gangal2019likelihood}. The baseline for the OSQ dataset is our implementation of \citet{zheng2020out}. 
%To compare OodGAN with baselines, we generated OOD data using OodGAN for both ROSTD and OSQ datasets and evaluated the performance of the OOD detection model. 
%We collected OOD data for both of these datasets with various settings of filtering mechanism. In the main results, we present only the best results out of all the filtering mechanism's settings we used for OodGAN. We used Soft Filter with maximal intent probability threshold set to 0.5 for ROSTD dataset, and Soft Filter with maximal intent probability threshold set to 0.25 for OSQ dataset. We present the experimental results and the impact of each filtering mechanism at the end of this section.

\autoref{tab:ROSTD-results} shows results on ROSTD dataset and \autoref{tab:OSQ-results} shows results on OSQ dataset. Results on ROSTD data are promising. They show around 65\% relative improvement in FPR 0.95 compared to baseline of our implementation of \citet{zheng2020out} and around 5\% absolute improvement in FPR 0.95 compared to baseline of \citet{gangal2019likelihood}. For the more challenging OSQ dataset, there is around 28\% relative improvement in both FPR 0.95 and FPR 0.90 compared to the baseline.

%Improvement in OSQ dataset is smaller due to a large number of intents. There are 150 intents whose distribution generator has to learn to generate examples belonging outside of them. Also, the generator depends on rewards from the auxiliary classifier. Thus, the auxiliary classifier's low performance can lead to a situation in which the generator generates IND data. Our analysis found that auxiliary classifier performance on the OSQ dataset is small on the part of intents, causing smaller improvements.

%OSQ dataset also contains training OOD examples. Thus we evaluated the performance of an OOD classifier trained on them as well in order to compare training and generated OOD examples. \autoref{tab:OSQ-results} shows that using training OOD data yields better AUROC and AUPR numbers compared to the model trained with generated data. However, our goal was to propose a system that can work with IND training examples only. Hence the comparison between classifier using training and generated OOD examples is not fair.

\begin{table}[]
\resizebox{\columnwidth}{!}{%
\begin{tabular}{@{}l|lllll@{}}
\hline
\multicolumn{1}{r|}{\begin{tabular}[c]{@{}r@{}}ROSTD \\ \citep{gangal2019likelihood}\end{tabular}} & AUROC $\uparrow$          & AUPR $\uparrow$           & \begin{tabular}[c]{@{}r@{}} FPR \\ 0.95\end{tabular} $\downarrow$       & \begin{tabular}[c]{@{}r@{}} FPR \\ 0.90\end{tabular} $\downarrow$        & \begin{tabular}[c]{@{}l@{}}IND \\ Acc.\end{tabular} $\uparrow$ \\ \hline
w.o. OOD                     & 97.64          & 93.86          & 8.10           & 5.56           & \textbf{99.05}                                               \\
\begin{tabular}[c]{@{}l@{}}Our implementation \\of \citet{zheng2020out}  \end{tabular}            & 88.67          & 54.84 & 37.82          & 26.04                & 88.00                                                             \\
\citet{gangal2019likelihood}              & 98.22          & \textbf{96.47} & 7.41          & -                & -                                                             \\
OodGAN                     & \textbf{98.99} & 96.26          & \textbf{2.59} & \textbf{1.37} & 98.31                                                        \\ \hline
\end{tabular}
}
\caption{OOD detection performance on the ROSTD dataset}
\label{tab:ROSTD-results}
\end{table}

\begin{table}[]
\resizebox{\columnwidth}{!}{%
\begin{tabular}{@{}l|lllll@{}}
\hline
\multicolumn{1}{r|}{\begin{tabular}[c]{@{}r@{}}OSQ \\ \citep{larson2019evaluation}\end{tabular}} & AUROC $\uparrow$           & AUPR $\uparrow$           & \begin{tabular}[c]{@{}r@{}} FPR \\ 0.95\end{tabular} $\downarrow$       & \begin{tabular}[c]{@{}r@{}} FPR \\ 0.90 \end{tabular} $\downarrow$       & \begin{tabular}[c]{@{}l@{}}IND \\ Acc.\end{tabular} $\uparrow$ \\ \hline
w.o. OOD                   & 90.89          & \textbf{97.99}          & 28.11          & 20.98          & 89.04                                                        \\
%Training OOD               & \textbf{91.72} & \textbf{98.29} & 31.78          & 21.87          & 86.18                                                        \\
\begin{tabular}[c]{@{}l@{}}Our implementation \\ of \citet{zheng2020out}\end{tabular}              & 88.79          & 58.22 & 36.49          & 26.87                & 88.00                                                             \\
OodGAN                   & \textbf{91.24}          & 97.79          & \textbf{26.07} & \textbf{19.29} & \textbf{90.11}                                               \\ \hline
\end{tabular}%
}
\caption{OOD detection performance on the OSQ dataset}
\label{tab:OSQ-results}
\end{table}

\begin{figure}[h!]%
    \centering
    \subfloat[\centering Model trained with no OOD]{{\includegraphics[width=\columnwidth]{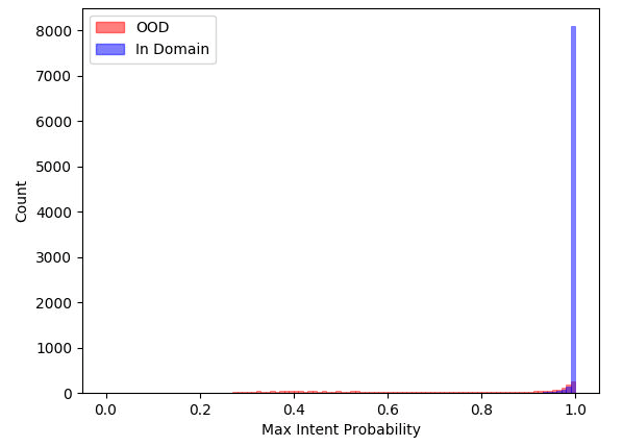} }}%
    \qquad
    \subfloat[\centering Model trained with generated OOD]{{\includegraphics[width=\columnwidth]{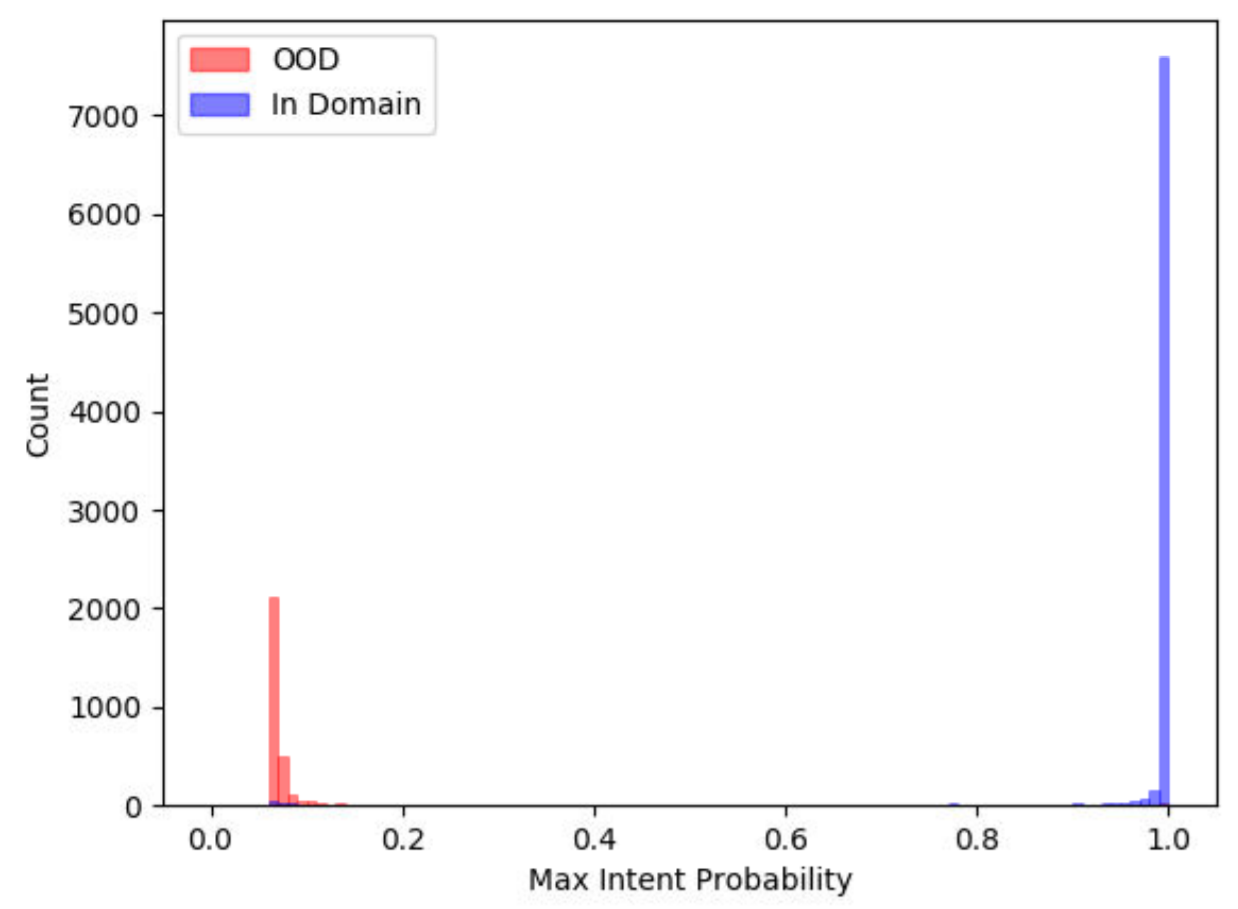} }}%
    \caption{Distributions of detection scores corresponding to the IND and OOD examples of the ROSTD dataset}%
    \label{ROSTD-histogram}%
\end{figure}

%To evaluate whether OodGAN helps the OOD recognition model to discriminate between OOD and IND examples, we plotted the histogram of the test data's maximum intent probability. We also plotted a histogram for the system trained without any OOD examples to see the improvement we gain if we use OodGAN.  Figures \ref{ROSTD-histogram} and \ref{OSQ-histogram} show the histogram for ROSTD and OSQ dataset respectively.
To evaluate whether OodGAN helps the threshold-based OOD detection model to discriminate between OOD and IND examples, we plotted the histogram of the test data's maximum intent probability for system trained with and without generated OOD examples. Figure \ref{ROSTD-histogram} shows the histogram for ROSTD dataset.
Probability scores for IND (blue) and OOD (red) data are spread out over all probability values when there are no OOD data used for model training. Thus it is hard to select a well discriminating threshold. The result of the model trained with OOD data is significantly better. 
%ROSTD dataset shows a clear separation between IND and OOD data, with IND data receiving high intent score and OOD data receiving low score. Graph for OSQ dataset is a little more spread out, but there is still a clear separation between IND and OOD data. Thus, OodGAN generates OOD examples that improve the OOD detection model's ability to discriminate between OOD and IND examples.
The graph shows a clear separation between IND and OOD data, with IND data receiving high intent score and OOD data receiving a low score.

%\begin{figure}[h]%
%    \centering
%    \subfloat[\centering Model trained with no OOD]{{\includegraphics[width=\columnwidth]{img/ROSTDhistbefore.PNG} }}%
%    \qquad
%    \subfloat[\centering Model trained with generated OOD]{{\includegraphics[width=\columnwidth]{img/ROSTD.PNG} }}%
%    \caption{Distributions of detection scores corresponding to the IND and OOD examples of the ROSTD dataset}%
%    \label{ROSTD-histogram}%
%\end{figure}

%\begin{figure}[h]%
%    \centering
%    \subfloat[\centering Model trained with no OOD]{{\includegraphics[width=\columnwidth]{img/OSQbefore.PNG} }}%
%    \qquad
%    \subfloat[\centering Model trained with generated OOD]{{\includegraphics[width=\columnwidth]{img/OSQ.PNG} }}%
%    \caption{Distributions of detection scores corresponding to the IND and OOD examples of the OSQ dataset}%
%    \label{OSQ-histogram}%
%\end{figure} 

\begin{figure}[h]
\centering
\includegraphics[width=\columnwidth]{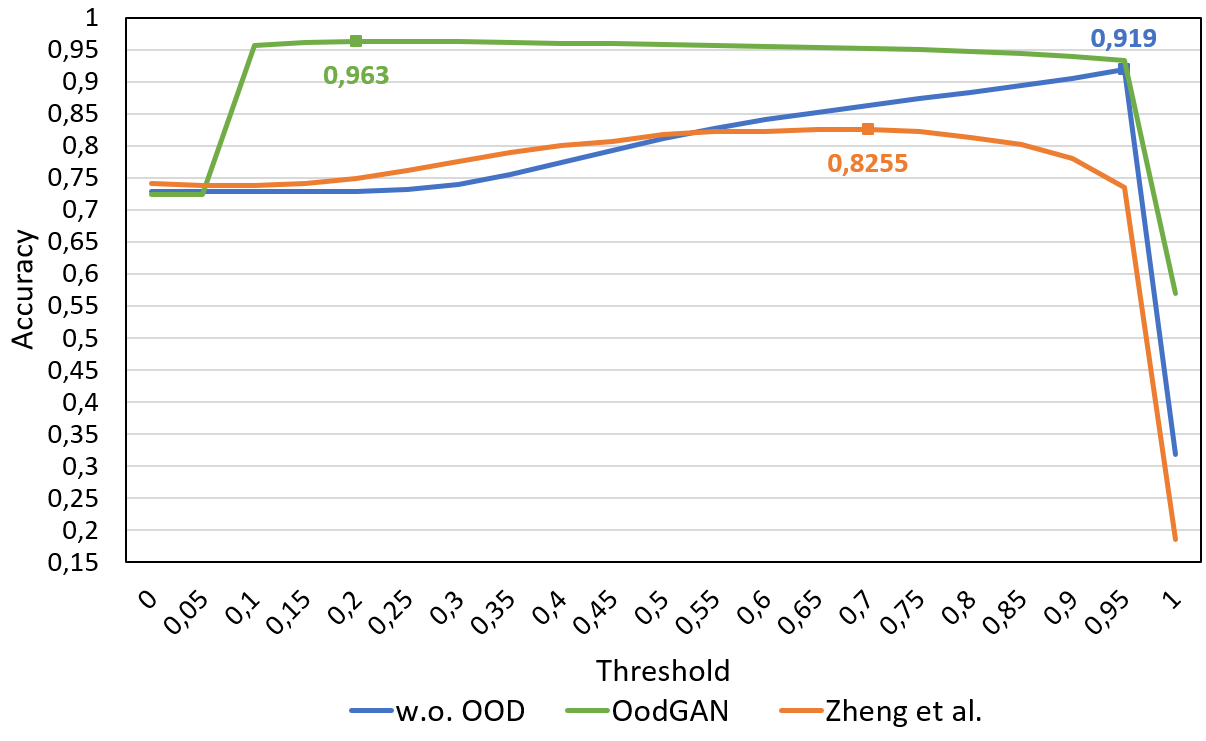}
\caption{Joint accuracy for ROSTD data across different threshold value. Points mark the highest joint accuracy of OOD detection and IND intent recognition.}
\label{ROSTD-thresholds}
\end{figure}

The OOD detection model is combined with IC in many real-world applications. For this reason, the joint accuracy of OOD detection and IND intent recognition is an important metric. 
We show how the joint accuracy depends on the selected threshold in Figure \ref{ROSTD-thresholds}.  
%We show how the joint accuracy depends on the selected threshold in Figures \ref{ROSTD-thresholds} and \ref{OSQ-thresholds}.  
To draw this graph, we select different thresholds, and we tag examples having an intent score below the threshold as OOD. We classify the intent for the rest.
Our proposed approach leads to high joint accuracy of OOD detection and IND intent recognition with low threshold values.
%Both OSQ and ROSTD datasets achieve high joint accuracy with low threshold values. 
That confirms that models trained with generated OOD assign low scores to OOD and high scores to IND examples.

%\begin{figure}[h]
%\centering
%\includegraphics[width=\columnwidth]{img/ROSTD_thresholds.png}
%\caption{Joint accuracy for ROSTD data across different threshold value}
%\label{ROSTD-thresholds}
%\end{figure}

%\begin{figure}[h]
%\centering
%\includegraphics[width=\columnwidth]{img/OSQ_thresholds.png}
%\caption{Joint accuracy for OSQ data across different threshold value}
%\label{OSQ-thresholds}
%\end{figure}

\begin{figure}[h]
\centering
\includegraphics[width=\columnwidth]{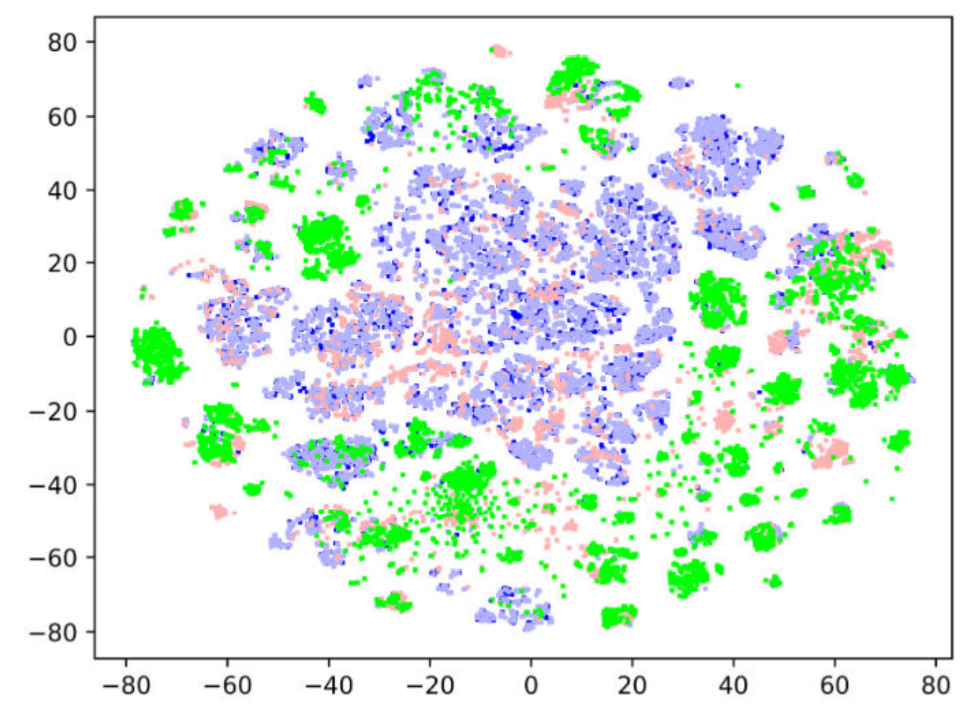}
\caption{t-SNE visualization of the BERT feature vectors associated with the examples from the ROSTD dataset. 
%The blue color represents IND examples, and the red color represents OOD examples.  A lighter shade of the color represents training examples, and a darker shade of the color represents validation and testing examples. 
IND examples are blue, testing OOD examples are red, and examples generated by OodGAN are green.}
\label{ROSTD-tsne}
\end{figure}

The separation between generated OOD examples and IND examples is visible in t-SNE \citep{hinton2002stochastic} visualization as well. %Figures \ref{ROSTD-tsne} and \ref{OSQ-tsne} show the t-SNE visualization of IND, golden OOD, and generated OOD data. 
Figure \ref{ROSTD-tsne} shows the t-SNE visualization of IND and generated OOD data. 
We can notice that generated data create recognizable clusters close to IND data but do not mix with it. Finally, we list OOD examples generated by OodGAN in table \ref{tab:ROSTD-examples}. 
%Finally, we list OOD examples generated by OodGAN in tables \ref{tab:ROSTD-examples} and \ref{tab:OSQ-examples}. It is clear from these examples that OodGAN can generate grammatically correct sentences, and it can also generate entities (such as "Game of Thrones" or "Galway"). This is another confirmation that examples generated by OodGAN are close to IND data. The results of t-SNE visualization and observation of generated OOD examples confirm the findings of \citet{zheng2020out} that training OOD recognition model with OOD data that are just outside IND distribution should be sufficient to handle most of the OOD requests during runtime.

%\begin{figure}[h]
%\centering
%\includegraphics[width=\columnwidth]{img/ROSTD Tsne.PNG}
%\caption{t-SNE visualization of the BERT feature vectors associated with the examples from the ROSTD dataset. The blue color represents IND examples, and the red color represents OOD examples. A lighter shade of the color represents training examples, and a darker shade of the color represents validation and testing examples. The green color represents examples generated by OodGAN.}
%\label{ROSTD-tsne}
%\end{figure}

%\begin{figure}[h]
%\centering
%\includegraphics[width=\columnwidth]{img/OSQ_Tsne.PNG}
%\caption{t-SNE visualization of the BERT feature vectors associated with the examples from the OSQ dataset. The blue color represents IND examples, and the red color represents OOD examples. The lighter shade of the color represents training examples, and the darker shade of the color represents validation and testing examples. The green color represents examples generated by OodGAN.}
%\label{OSQ-tsne}
%\end{figure}

\begin{table}[h]
\resizebox{\columnwidth}{!}{%
\begin{tabular}{@{}lllll@{}}
\hline
IND Examples         & \begin{tabular}[c]{@{}l@{}}Should I be expecting rain today\\ I need a new alarm for 8:30 am\\ Show my reminders\\ Show me the extended forecast please\\ Snooze alarm for 5 more minutes\end{tabular}                 \\ \hline
OOD Examples         & \begin{tabular}[c]{@{}l@{}}Why do people watch television\\ Where do pineapples grow\\ Should I go to the mall today or tomorrow\\ Tell me how to install a pool\\ Transfer my PayPal balance to my bank\end{tabular}  \\ \hline
\begin{tabular}[c]{@{}l@{}}Generated by \\ OodGAN \end{tabular} & \begin{tabular}[c]{@{}l@{}}Remind me of my 4pm and Game of Thrones alarm\\ When should I unpack\\ Add day at workout please\\ Give me my Sarasota appointment\\ Do I need to pack to Galway this umbrella\end{tabular} \\ \hline
\end{tabular}
}
\caption{
Examples sampled from the IND and OOD test set of the ROSTD dataset and OOD utterances generated using OodGAN model.
}
\label{tab:ROSTD-examples}
\end{table}

\section{Conclusion}
This paper proposed a novel OOD data generation model OodGAN that generates OOD examples that improved OOD detection performance in a dialog system. The model does not require any OOD training examples. Moreover, the model does not rely on the auto-encoder to map utterances into latent space, reducing the model size. It models the data generator as a stochastic policy in reinforcement learning instead. The model uses two rewards for the generator. The discriminator's reward guides the generator to generate examples as close to the IND data as possible. The auxiliary intent classifier's reward guides the generator to generate examples with low probabilities for all intent classes.
%The model uses two rewards for the generator. The first reward originates from the discriminator. Its role is to make generated examples as close to the IND data as possible. The second reward originates from an auxiliary intent classifier pre-trained to classify the intent of IND examples. The generator receives the highest reward when the auxiliary intent classifier assigns a generated example with low probabilities for all intent classes. 
%We also propose a Filtering Mechanism responsible for removing generated examples that are equal or too close to the IND distribution. Our experiments show that OOD examples generated by OodGAN and post-processed by filtering mechanism improve the performance of the OOD detection problem.
Our experiments show that OOD examples generated by OodGAN improve the performance of the OOD detection problem.

\bibliography{anthology,custom}

\begin{thebibliography}{16}
\expandafter\ifx\csname natexlab\endcsname\relax\def\natexlab#1{#1}\fi

\bibitem[{Donahue and Rumshisky(2018)}]{donahue2018adversarial}
David Donahue and Anna Rumshisky. 2018.
\newblock Adversarial text generation without reinforcement learning.
\newblock \emph{arXiv preprint arXiv:1810.06640}.

\bibitem[{Gangal et~al.(2019)Gangal, Arora, Einolghozati, and
  Gupta}]{gangal2019likelihood}
Varun Gangal, Abhinav Arora, Arash Einolghozati, and Sonal Gupta. 2019.
\newblock Likelihood ratios and generative classifiers for unsupervised
  out-of-domain detection in task oriented dialog.
\newblock \emph{arXiv preprint arXiv:1912.12800}.

\bibitem[{Guo et~al.(2017)Guo, Pleiss, Sun, and
  Weinberger}]{guo2017calibration}
Chuan Guo, Geoff Pleiss, Yu~Sun, and Kilian~Q Weinberger. 2017.
\newblock On calibration of modern neural networks.
\newblock \emph{arXiv preprint arXiv:1706.04599}.

\bibitem[{Hendrycks and Gimpel(2017)}]{Hendrycks2017ABF}
Dan Hendrycks and Kevin Gimpel. 2017.
\newblock A baseline for detecting misclassified and out-of-distribution
  examples in neural networks.
\newblock \emph{ArXiv}, abs/1610.02136.

\bibitem[{Hendrycks et~al.(2019)Hendrycks, Mazeika, and
  Dietterich}]{Hendrycks2019DeepAD}
Dan Hendrycks, Mantas Mazeika, and Thomas~G. Dietterich. 2019.
\newblock Deep anomaly detection with outlier exposure.
\newblock \emph{ArXiv}, abs/1812.04606.

\bibitem[{Hinton and Roweis(2002)}]{hinton2002stochastic}
Geoffrey~E Hinton and Sam Roweis. 2002.
\newblock Stochastic neighbor embedding.
\newblock \emph{Advances in neural information processing systems},
  15:857--864.

\bibitem[{Joulin et~al.(2016)Joulin, Grave, Bojanowski, and
  Mikolov}]{joulin2016bag}
Armand Joulin, Edouard Grave, Piotr Bojanowski, and Tomas Mikolov. 2016.
\newblock Bag of tricks for efficient text classification.
\newblock \emph{arXiv preprint arXiv:1607.01759}.

\bibitem[{Kim(2014)}]{kim2014convolutional}
Yoon Kim. 2014.
\newblock Convolutional neural networks for sentence classification.
\newblock \emph{arXiv preprint arXiv:1408.5882}.

\bibitem[{Lakshminarayanan et~al.(2017)Lakshminarayanan, Pritzel, and
  Blundell}]{lakshminarayanan2017simple}
Balaji Lakshminarayanan, Alexander Pritzel, and Charles Blundell. 2017.
\newblock Simple and scalable predictive uncertainty estimation using deep
  ensembles.
\newblock In \emph{Advances in neural information processing systems}, pages
  6402--6413.

\bibitem[{Larson et~al.(2019)Larson, Mahendran, Peper, Clarke, Lee, Hill,
  Kummerfeld, Leach, Laurenzano, Tang et~al.}]{larson2019evaluation}
Stefan Larson, Anish Mahendran, Joseph~J Peper, Christopher Clarke, Andrew Lee,
  Parker Hill, Jonathan~K Kummerfeld, Kevin Leach, Michael~A Laurenzano,
  Lingjia Tang, et~al. 2019.
\newblock An evaluation dataset for intent classification and out-of-scope
  prediction.
\newblock \emph{arXiv preprint arXiv:1909.02027}.

\bibitem[{Lee and Shalyminov(2019)}]{lee2019contextual}
Sungjin Lee and Igor Shalyminov. 2019.
\newblock Contextual out-of-domain utterance handling with counterfeit data
  augmentation.
\newblock In \emph{ICASSP 2019-2019 IEEE International Conference on Acoustics,
  Speech and Signal Processing (ICASSP)}, pages 7205--7209. IEEE.

\bibitem[{Ren et~al.(2019)Ren, Liu, Fertig, Snoek, Poplin, DePristo, Dillon,
  and Lakshminarayanan}]{Ren2019LikelihoodRF}
J.~Ren, Peter~J. Liu, E.~Fertig, Jasper Snoek, Ryan Poplin, Mark~A. DePristo,
  Joshua~V. Dillon, and Balaji Lakshminarayanan. 2019.
\newblock Likelihood ratios for out-of-distribution detection.
\newblock In \emph{NeurIPS}.

\bibitem[{Ryu et~al.(2018)Ryu, Koo, Yu, and Lee}]{ryu2018out}
Seonghan Ryu, Sangjun Koo, Hwanjo Yu, and Gary~Geunbae Lee. 2018.
\newblock Out-of-domain detection based on generative adversarial network.
\newblock In \emph{Proceedings of the 2018 Conference on Empirical Methods in
  Natural Language Processing}, pages 714--718.

\bibitem[{Williams(1992)}]{williams1992simple}
Ronald~J Williams. 1992.
\newblock Simple statistical gradient-following algorithms for connectionist
  reinforcement learning.
\newblock \emph{Machine learning}, 8(3-4):229--256.

\bibitem[{Yu et~al.(2017)Yu, Zhang, Wang, and Yu}]{yu2017seqgan}
Lantao Yu, Weinan Zhang, Jun Wang, and Yong Yu. 2017.
\newblock Seqgan: Sequence generative adversarial nets with policy gradient.
\newblock In \emph{Thirty-first AAAI conference on artificial intelligence}.

\bibitem[{Zheng et~al.(2020)Zheng, Chen, and Huang}]{zheng2020out}
Yinhe Zheng, Guanyi Chen, and Minlie Huang. 2020.
\newblock Out-of-domain detection for natural language understanding in dialog
  systems.
\newblock \emph{IEEE/ACM Transactions on Audio, Speech, and Language
  Processing}, 28:1198--1209.

\end{thebibliography}
\bibliographystyle{acl_natbib}

%\appendix
%
%\section{Example Appendix}
%\label{sec:appendix}
%
%This is an appendix.

\end{document}